\documentclass{article}

\usepackage[nonatbib,final]{neurips_2020-2}

\usepackage[utf8]{inputenc} % allow utf-8 input
\usepackage[T1]{fontenc}    % use 8-bit T1 fonts
\usepackage{hyperref}       % hyperlinks
\usepackage{url}            % simple URL typesetting
\usepackage{booktabs}       % professional-quality tables
\usepackage{amsfonts}       % blackboard math symbols
\usepackage{nicefrac}       % compact symbols for 1/2, etc.
\usepackage{microtype}      % microtypography

\usepackage{graphicx}
\usepackage{amsmath}
\usepackage{gensymb}
\usepackage{placeins}
\usepackage{subcaption}
\usepackage[font=small,labelfont=bf]{caption}

\title{Unsupervised Domain Adaptation \\
for Visual Navigation}

\author{Shangda Li$^*$,~~ Devendra Singh Chaplot\thanks{equal contribution},~~ Yao-Hung Hubert Tsai,~~ Yue Wu,\\
    \textbf{Louis-Philippe Morency,~~ Ruslan Salakhutdinov}\\
	Carnegie Mellon University\\
	{ \texttt \tt iharryharryli@gmail.com,}\ {\texttt\tt \{chaplot,yaohungt,ywu5,morency,rsalakhu\}@andrew.cmu.edu}
}
\def\methodname{Policy-Based Image Translation}
\def\methodabbr{PBIT}

\begin{document}

\maketitle

\begin{abstract}
Advances in visual navigation methods have led to intelligent embodied navigation agents capable of learning meaningful representations from raw RGB images and perform a wide variety of tasks involving structural and semantic reasoning. However, most learning-based navigation policies are trained and tested in simulation environments. In order for these policies to be practically useful, they need to be transferred to the real-world. 
In this paper, we propose an unsupervised domain adaptation method for visual navigation. Our method translates the images in the target domain to the source domain such that the translation is consistent with the representations learned by the navigation policy. The proposed method outperforms several baselines across two different navigation tasks in simulation. We further show that our method can be used to transfer the navigation policies learned in simulation to the real world.
\end{abstract}

\section{Introduction}
\label{sec:intro}

In the past few years, a lot of progress has been made in learning to navigate from first-person RGB images. Reinforcement learning have been applied to train navigation policies to navigate to goals according to coordinates~\cite{gupta2017cognitive,ans,wijmans2019decentralized}, images~\cite{zhu2017target}, object labels~\cite{gupta2017cognitive,yang2018visual}, room labels~\cite{wu2018building,wu2019bayesian} and language instructions~\cite{hermann2017grounded,chaplot2018gated,anderson2018vision,fried2018speaker,chen2019touchdown,wang2019reinforced}. However, such navigation policies are predominantly trained and tested in simulation environments. Our goal is to have such navigation capabilities in the real-world. While some progress has been made towards moving from game-like simulation environments to more realistic simulation environments based on reconstructions~\cite{gibsonenv,Matterport3D,replica19arxiv} or 3D modeling~\cite{ai2thor}, there is still a significant gap between simulation environments and real-world.

Training the above navigation policies in the real-world has not been possible as current reinforcement learning methods typically require millions of samples for training. Even if we parallelize the training across multiple robots, it will still require multiple weeks on training with constant human supervision due to safety concerns and battery limitations. This makes real-world training practically infeasible and leaves us with the other option of transferring models trained in simulation to the real-world, which highlights the importance of domain adaptation methods.

Among domain adaptation techniques, unsupervised methods are favorable because it is extremely expensive to collect parallel data for the purpose of visual navigation. It essentially requires reconstructing real-world scenes in the simulator separately for all possible scenarios one might deploy the navigation model in such as different lighting conditions, time of day, indoor vs outdoor, weather conditions, and so on. Reconstructing real-world scenes is a tedious job requiring specialized cameras and significant human effort. Unsupervised learning methods have the potential to overcome this difficulty since they require only a few real-world images taken by regular cameras.

One possible solution involves using unsupervised image translation techniques to translate visual perception from simulation to real-world and adapt the navigation policy learned in simulation to the real-world. Although there already exists a rich amount of prior work in unsupervised image translation techniques that transfer images from one domain to another~\cite{zhu2017unpaired,liu2017unsupervised,MUNIT}, prior techniques are not well suited for navigation since the image translations are agnostic of the navigation policy and instead focus on photo-realisticity and clarity. 

\begin{figure}[t]
\centering
%\vspace{-5mm}
\includegraphics[width=\textwidth]{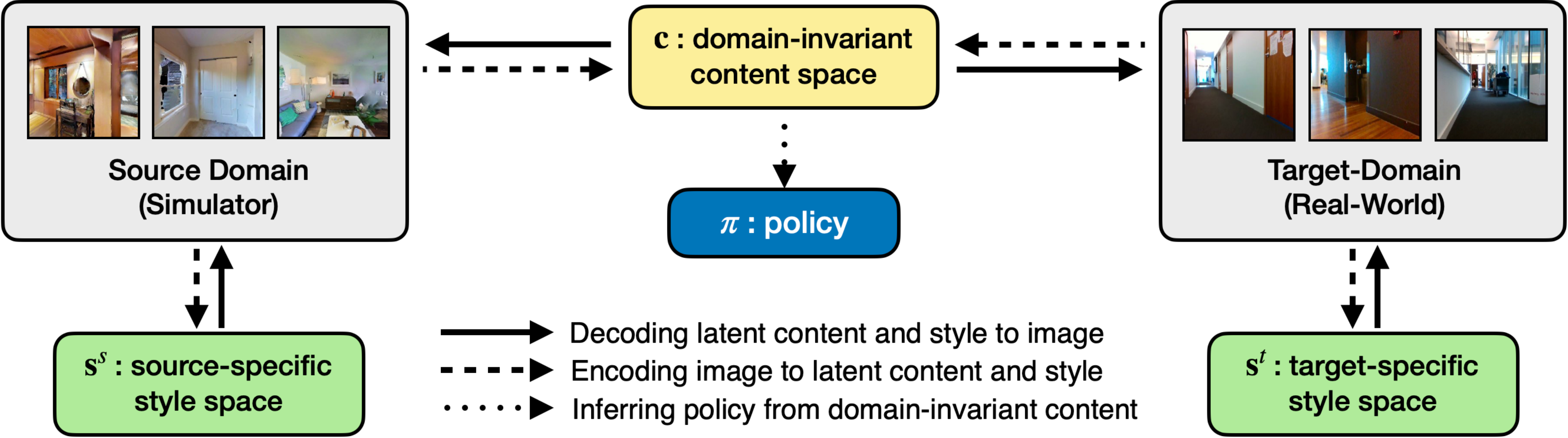}
%\vspace{-2.5mm}
\caption{\small {\bf PBIT.} The proposed policy-based image translation for unsupervised visual navigation adaptation.}
\vspace{-4mm}
\label{fig:overall}
\end{figure}

In this paper, we propose an unsupervised domain adaptation method for transferring navigation policies from simulation to the real-world, by unsupervised image translation subject to the constraint that the image translation respects agent's policy. 
%The key idea behind our method is a policy-based image translation. 
% We translate images across domains such that the representations learnt by the navigation policy are consistent across domains. 
In order to learn policy-based image translation (PBIT) in an unsupervised fashion, we devise a disentanglement of content and style in images such that the representations learnt by the navigation policy are consistent for images with the same content with different styles. See Figure~\ref{fig:overall} for the illustration of PBIT. Our experiments show that the proposed method outperforms the baselines in transferring navigation policies for different tasks between two simulation domains and from simulation to the real-world.

\section{Related Work}
\label{sec:related}

Simulation to real-world (Sim-to-Real) transfer of visual navigation policies requires the adaptation for both visual perception and agent's policy. Among its wide range of relevant literature, we focus on discussing related work on {\em visual navigation}, {\em visual domain adaptation} and {\em policy transfer}.

{\bf Visual Navigation.} Visual Navigation has been widely studied in robotics for over two decades. Bonin-Font et al.~\cite{bonin2008visual} provide an in-depth survey of visual navigation methods in classical robotics. In the past few years, there has been an increasing focus on learning-based methods for visual navigation. This includes methods which tackle navigation tasks primarily requiring geometrical scene understanding such as the \textit{pointgoal} task~\cite{gupta2017cognitive, anderson2018evaluation} where the relative coordinate to the goal is given and the \textit{exploration} task~\cite{chen2019learning,fang2019scene,ans} where the objective is to maximize the explored area. There has also been a lot of work on navigation tasks involving more semantics such as image-goal~\cite{zhu2017target,chaplot2020neural}, object-goal~\cite{yang2018visual,wortsman2019learning,mousavian2019visual,chaplot2020object,chang2020semantic}, high-level language goal~\cite{hermann2017grounded,chaplot2018gated} and low-level language instructions~\cite{anderson2018vision,fried2018speaker}.

While performance on semantic navigation tasks is still far from perfect even in simulation, recent improvements in both visual simulation quality~\cite{savva2019habitat,gibsonenv,Matterport3D,replica19arxiv} as well as algorithms~\cite{sax2019learning,ans,wijmans2019decentralized} have led to impressive results on geometric navigation tasks. However, most of the above works train navigation policies using reinforcement or imitation learning in simulation and test on different scenes in the same domain in the simulator. Some prior works which tackle sim-to-real transfer for navigation policies directly transfer the policy trained in simulation to the real-world without any domain adaptation technique~\cite{gupta2017cognitive,ans}. We show that the proposed domain adaptation method can lead to large improvements over direct policy transfer. 

{\bf Visual Domain Adaptation.} Simulation and real-world can be viewed as two distinct visual domains, and adapting their visual perceptions can be regarded as an image-to-image translation task. Thanks to the success of Generative Adversarial Networks (GANs)~\cite{goodfellow2014generative} for matching cross-domain distribution, we are able to adapt an image across domains without changing its context. For example, pix2pix~\cite{isola2017image} changes only the style of an image (e.g., photograph $\rightarrow$ portrait) while preserving its context (e.g., the same face of a person). We note that, for Sim2Real navigation, some amount of context should be preserved across domains, such as the obstacles and walls, to prevent collisions.

If we have access to the paired cross-domain images, then pix2pix~\cite{isola2017image} and BicycleGAN~\cite{zhu2017toward} serve as good candidates to model the context-preserving adaptation. However, the paired data between simulation and real-world is notoriously hard to collect~\cite{tzeng2015adapting} or even do not exist (e.g., we cannot always build simulators for new environments). To tackle this challenge, numerous visual domain adaptation approaches~\cite{taigman2016unsupervised,shrivastava2017learning,zhu2017unpaired,kim2017learning,yi2017dualgan,liu2017unsupervised} have been proposed to relax the constraint of requiring paired data during training time. Nevertheless, the above methods still assume one-to-one correspondence across domains. As an example, these models can only generate the same target-domain image given a source-domain image. We argue that it is more realistic to assume a many-to-many mapping between simulation and real-world.

To learn multimodal mappings without paired data, prior works~\cite{MUNIT, lee2018diverse} disentangle the context and style of an image. Precisely, they assume the context is shared across domains and the styles are specific to each domain. Note that these models focus on realistic image generation, and hence it remains unclear on how image translation benefits cross-domain visual navigation. To further bridge the gap between navigation and image translation, our key idea is to ensure the agent's navigation policy be consistent under domain translation. As a consequence, we propose to enforce constraints such that the agent's policy is only inferred from the shared context across simulation and real-world. 

{\bf Policy Transfer.} Existing works on sim-to-sim or sim-to-real policy transfer require domain knowledge specific to certain environments and tasks and extra human supervision. \cite{8620258} employs semantic loss and shift loss for consistent generation to facilitate the domain translation. \cite{Sadeghi2019DIViSDI} build a new simulator with diverse layouts populated with diverse furniture placements and defines several auxiliary tasks that are specific to the Collision-Free Goal Reaching task. \cite{Mller2018DrivingPT} trains the domain translation module of its driving agent on human-labeled segmentation dataset to generalize its learned policy from one domain to the other. \cite{Bousmalis2018UsingSA} utilizes customized simulator to improve a policy's real-world performance, given a policy already trained in the real world, which is different from our setting where the policy is trained in the simulation and then gets transferred to the real-world. \cite{Gordon2019SplitNetSA} targets similar visual navigation problems and defines auxiliary tasks for better sim-to-sim transfer, which is complementary to our contribution. \cite{Blukis2019LearningTM} relies on Supervised Learning for Visitation Prediction which requires human-supervised trajectories of both the real world and the simulation.

%{\bf Sim2Real Robot Learning.} Some prior works have tackled sim2real transfer for various robot learning tasks. Tobin et al.~\cite{tobin2017domain} propose a technique of domain randomization for a manipulation task, which involves randomizing the textures of objects during training. While this technique can help in sim2real transfer for table-top manipulation settings, it is not well suited for navigation as it real-world environments involve a wide diversity of potentially unseen objects as well as different relative configurations of objects and clutter.

%{\color{red} [this part is for Devendra] perhaps briefly discussing what other kind of Sim2Real tasks other than visual navigation. Like robot arms. I listed what I knew below.}

%VG-Goggles~\cite{zhu2017unpaired}

%Domain alignment with weakly supervised pairwise constraints~\cite{tzeng2015adapting}

%Invariant Feature Transfer~\cite{gupta2017learning}

%Domain Randomnization~\cite{tobin2017domain}

%Progressive Nerual Network~\cite{rusu2016sim}

%Inverse Dynamic Model~\cite{christiano2016transfer}

\section{Methods}
%\label{sec:task}
\vspace{-4pt}

Denote the source domain as $(\mathcal{S}^s, \mathcal{A}, P^s)$ and the target domain as $(\mathcal{S}^t, \mathcal{A}, P^t)$. $\mathcal{S}$ is the state space, $\mathcal{A}$ is the set of actions, and $P:\mathcal{S}\times \mathcal{A}\times \mathcal{S}\to \mathbb{R}$ is the transition probability distribution. Note that we assume action spaces $\mathcal{A}$ are shared across domains. Let $\pi^s:\mathcal{S}^s \to \mathcal{A}$ be a navigation policy in the source domain $s$ (the navigation policy is given). For our task setup, we have access to some target-domain images $I^t \in \mathcal{S}^t$  during training, but we cannot perform target-domain policy ($\pi^t$) training. Our objective is to learn a many-to-many mapping $F:\mathcal{S}^t\to \mathcal{S}^s$ such that the navigation policy under the source-to-target mapping, $\pi^t(I^t) = \pi^s(F(I^t))$, is effective in the target domain $t$. Under Sim2Real setting, the source domain refers to simulator and the target domain refers to real-world. Unless specified otherwise, we abbreviate $\pi^s$ as $\pi$ for the rest of the paper. 

%The remaining part of this section shall describe our proposed `\methodname' (\methodabbr). 

%The correlation between the two environments are defined by a many-to-many function $F:\mathcal{P}(S_s)\to \mathcal{P}(S_t)$ such that there exists optimal policy $\pi_s: \mathcal{S}_s \to \mathcal{A}$, $\pi_t: \mathcal{S}_t \to \mathcal{A}$, $\pi_s(S_s) = \pi_t(F(S_s))$.
\begin{figure}
    \centering
    \vspace{4mm}
    \includegraphics[width=0.99\textwidth, angle=0]{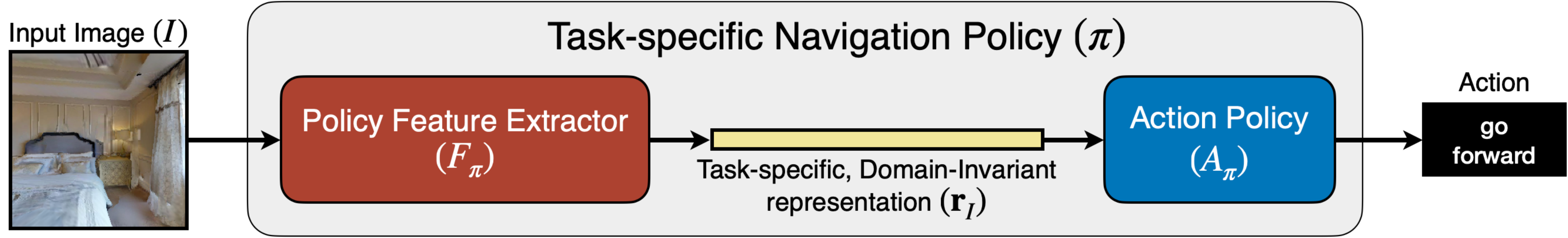}
    % \vspace{-3mm}
    \caption{\small \textbf{Policy Decomposition.} The Task-specific Navigation Policy ($\pi$) can be sequentially decomposed into a Policy Feature Extractor ($F_{\pi}$) and a Action Policy ($A_{\pi}$) such that $F_{\pi}$ extracts all task-specific features (${\bf r}_I$) and throws away all domain-specific features from the input image ($I$) and $A_{\pi}$ learns an action distribution function over the task-specific features.}
    \vspace{-6mm}
    \label{fig:policy_decomposition}
\end{figure}

\subsection{Policy Decomposition}
\vspace{-4pt}
As our objective is to transfer the task-specific navigation policy across domains, we assume that the task itself is domain-invariant. 
%This is true for most navigation tasks studied in the past literature{\color{red}~\cite{xxx}}. 
As a consequence, given a policy $\pi$ (for the navigation task in the source domain), we assume that some intermediate task-specific representation inferred by the policy is invariant from the source to the target domain. For example, a simple obstacle avoidance navigation policy would extract task-specific and domain-invariant features such as distance to obstacles at various angles and then learn a policy over these features. Let $\pi$ be sequentially decomposed into a Policy Feature Extractor ($F_{\pi}$) and a Action Policy ($A_{\pi}$). $F_{\pi}$ extracts all task-specific features (${\bf r}_I$ with $I$ indicating the input image) and throws away all domain-specific features in the input image. $A_{\pi}$ learns an action distribution function over the task-specific features. We illustrate the policy decomposition in Figure~\ref{fig:policy_decomposition}.

\subsection{Policy-based Consistency Loss}
\vspace{-4pt}
Recall that our objective is to learn an image translation model $F:\mathcal{S}^t\to \mathcal{S}^s$, such that $\pi^t(I^t) = \pi^s(F(I^t))$. Based on the policy decomposition described above, different translations of the same target-domain image would have similar task-specific features.
%By policy decomposition, given a target-domain image, different translated images to the source domain would have similar task-specific features. 
Precisely, if $I^{s}_1$ and $I^{s}_2$ are the translated images to the source domain from the same target-domain image $I^t$, then $F_{\pi}(I^{s}_1) \approx F_{\pi}(I^{s}_2)$.

\begin{figure}[t]
    % \vspace{-3mm}
    \centering
    % \vspace{-3mm}
    \includegraphics[width=0.99\textwidth, angle=0]{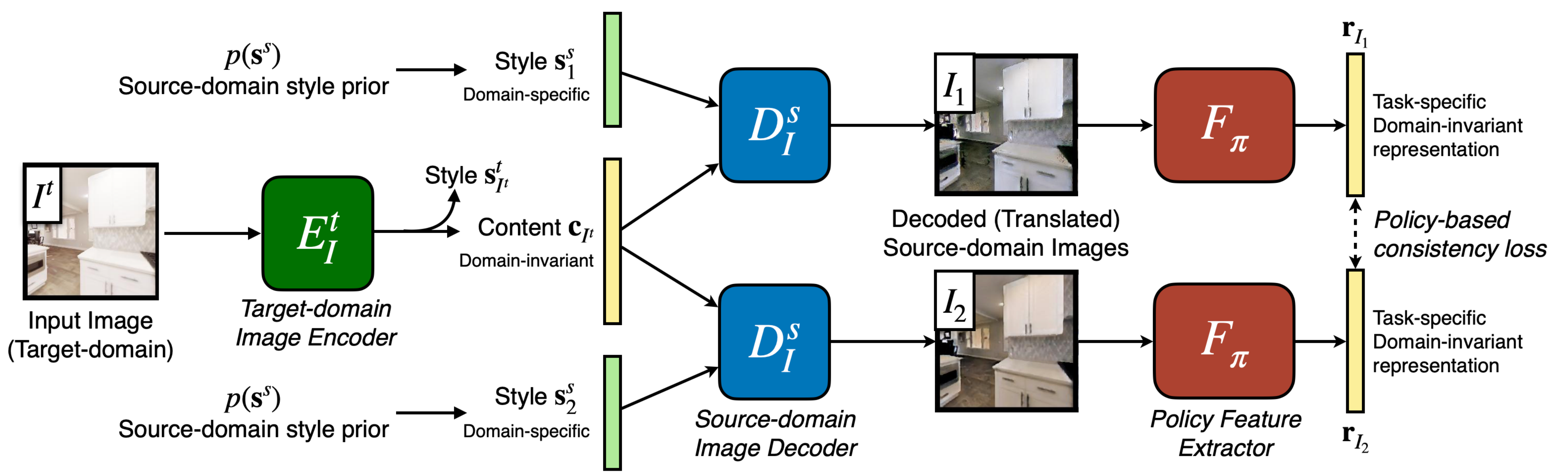}
    % \vspace{-3mm}
    \caption{\small \textbf{Policy-based Consistency Loss.} Since the task is domain-invariant, task-specific representations obtained from different domain-specific styles but the same domain-invariant content should be similar.}
    \vspace{-3mm}
    \label{fig:policy_loss}
\end{figure}

To achieve the above policy consistency
in an unsupervised fashion, we take inspiration from style and content-based unsupervised methods designed for image translation~\cite{MUNIT}. We assume that each image can be decomposed into a domain-invariant content representation (${\bf c}$) and a domain-specific style representation (${\bf s}$). Let $E_{I}^s$ be an Image Encoder for domain $s$ which encodes an image ($I$) to domain-invariant content (${\bf c}$) and domain-specific style (${\bf s}^s$): $
E_{I}^s(I) = ({\bf c}_I, {\bf s}^s_I)$. On the contrary, let $D_{I}^s$ be an Image Decoder which is the inverse of the Image Encoder: $D_{I}^s({\bf c}_I, {\bf s}^s_I) = I$.

Since we assume the navigation task is domain-invariant, all the the task-specific features are a subset of content representation ${\bf r}_I \in {\bf c}_I$. Therefore, images generated from different styles but same content should lead to the same task-specific features as shown in Figure~\ref{fig:policy_loss}. We operationalize this idea using the following policy-based consistency loss:
\vspace{4pt}
\begin{equation}
\small
\mathcal{L}_{pol} = \mathbb{E}_{{\bf c}_{I^t}: ({\bf c}_{I^t}, 
\_)\in E_{I}^t (I^t), I^t\sim \mathcal{S}^t, {\bf s}_{1}^s \sim p({\bf s}^s), {\bf s}_{2}^s \sim p({\bf s}^s)} [||F_{\pi}(D^s_I({\bf c}_{I^t},{\bf s}_{1}^s)) - F_{\pi}(D^s_I({\bf c}_{I^t},{\bf s}_{2}^s))||_1]
\label{eq:policy_consistency}
\end{equation}
\vspace{6pt}
with ${\bf s}_{1}^s$ and ${\bf s}_{2}^s$ being two distinct styles sampled from the the prior distribution $p({\bf s}^s) := \mathcal{N}({\bf 0}, {\bf I})$. %  a multivariate Gaussian with zero mean and diagonal unit covariance.

Note that in the above equation, $F_{\pi}(\cdot)$ is part of the given navigation policy. We assume the navigation policy is trained before deciding the target domain; hence, $F_{\pi}(\cdot)$ is fixed during the domain adaptation phase. This ensures that the presented policy-based image translation (PBIT) can be used for transferring a policy across domains (potentially not anticipated during policy training) without re-training the the navigation policy.

\subsection{Reconstruction and Adversarial Loss}
Using just policy-based consistency loss would make decoder $D_I$ ignore the style and decode based only on the content. Inspired by prior work~\cite{zhu2017unpaired,MUNIT}, to encourage the content to be domain-invariant and style representations to be domain-specific, we adopt the following image and latent representation reconstruction losses, and use $\mathcal{N}({\bf 0}, {\bf I})$ for the prior distributions of styles $p({\bf s}^s)$ and $p({\bf s}^t)$:
\begin{equation}
\small
\begin{split}
\mathcal{L}_{im\_rec} = & \mathbb{E}_{I^t\sim \mathcal{S}^t}[
||D_I^t(E_I^t(I^t)) - I^t ||_1] + 
\mathbb{E}_{I^s\sim \mathcal{S}^s}[
||D_I^s(E_I^s(I^s)) - I^s ||_1],\\
\mathcal{L}_{lat\_rec} = & \mathbb{E}_{{\bf c}_{I^t}:({\bf c}_{I^t}, \_)\in E_I^t(I^t), I^t\sim \mathcal{S}^t, 
{\bf s}^s \sim p({\bf s}^s )}[
||E_I^s(D_I^s({\bf c}_{I^t}, {\bf s}^s)) - ({\bf c}_{I^t}, {\bf s}^s) ||_1] \\ 
& +  \mathbb{E}_{{\bf c}_{I^s}:({\bf c}_{I^s}, \_)\in E_I^s(I^s), I^s\sim \mathcal{S}^s, 
{\bf s}^t \sim p({\bf s}^t )}[
||E_I^t(D_I^t({\bf c}_{I^s}, {\bf s}^t)) - ({\bf c}_{I^s}, {\bf s}^t) ||_1].
\end{split}
\label{eq:recon}
\end{equation}

We also use adversarial losses to match the distribution of images to their respective domains. Let ${\rm Dis}^{s}$ and ${\rm Dis}^{t}$ be the discriminators for the source and target domains:
\begin{equation}
\begin{split}
\mathcal{L}_{adv} = &
 \mathbb{E}_{{\bf c}_{I^t}:({\bf c}_{I^t}, \_)\in E_I^t(I^t), I^t\sim \mathcal{S}^t, 
{\bf s}^s \sim p({\bf s}^s )}[\log {\rm Dis}^{s}(D_I^s({\bf c}_{I^t}, {\bf s}^s))] \\ & +
\mathbb{E}_{{\bf c}_{I^s}:({\bf c}_{I^s}, \_)\in E_I^s(I^s), I^s\sim \mathcal{S}^s, 
{\bf s}^t \sim p({\bf s}^t )}[\log {\rm Dis}^t(D_I^t({\bf c}_{I^s}, {\bf s}^t))]\\ & + \mathbb{E}_{I^s\sim \mathcal{S}^s}[\log(1 - {\rm Dis}^s(I^s))]
+ \mathbb{E}_{I^t\sim \mathcal{S}^t}[\log(1 - {\rm Dis}^t(I^t))].
\end{split}
\label{eq:adver}
\end{equation}
Putting everything together, our overall objective is 
\begin{equation}
\mathcal{L}_{full} := \lambda_{pol} \mathcal{L}_{pol} 
+ \lambda_{im\_rec}
\mathcal{L}_{im\_rec} + \lambda_{lat\_rec}\mathcal{L}_{lat\_rec} + \lambda_{adv} \mathcal{L}_{adv}, 
\label{eq:full_objective}
\end{equation}
where $\lambda$s are hyper-parameters controlling the weight of each loss during training.
The cross-domain image translation model consists of $D_I^t, E_I^t, D_I^s$ and $E_I^s$ as described above and the optimization admits a mix-max objective:
\vspace{6pt}
\begin{equation}
D_I^t, E_I^t, D_I^s, E_I^s = \arg\min_{D_I^t, E_I^t, D_I^s, E_I^s}\,\,\max_{{\rm Dis}^t, {\rm Dis}^s} \mathcal{L}_{full}.
\label{eq:minmax}
\end{equation}

\section{Experimental Setup}
\label{sec:setup}
We conduct two sets of experiments to test the domain adaptation of navigation policies in Sim-to-Sim and Sim-to-Real settings. In the Sim-to-Sim experiments, we adapt navigation policies trained in the Gibson~\cite{gibsonenv} domain to the Replica~\cite{replica19arxiv} domain in the Habitat Simulator~\cite{savva2019habitat}. For Sim-to-Real, we adapt policies trained in Gibson to real-world office scenes. For training the domain adaptation models, we collect 7200 images randomly in 72 training scenes in Gibson, 7200 images in 18 test scenes in Replica, and 1125 images in the real-world. 

We study domain adaptation for two navigation tasks, PointGoal and Exploration for our experiments. The PointGoal task~\cite{anderson2018vision, savva2019habitat} involves navigating to the target location specified using point coordinates. Success Rate and Success weighted by Path Length (SPL)~\cite{anderson2018vision} as evaluation metrics for PointGoal. An episode is considered successful if the agent is within $0.2m$ radius of the goal location at the end of the episode. The Exploration task~\cite{chen2019learning,ans} involves maximizing the coverage or the explored area within a fixed time budget of 500 steps. A traversable point is defined to be explored by the agent if it is in the field-of-view of the agent and less than $3m$ away from the agent. We use two evaluation metrics, the absolute coverage area in $m^2$ (Explored Area) and the proportion of area explored in the scene (Explored Ratio).

For both the tasks, the agent has two sensors: RGB camera ($3\times256\times256$) and an odometry sensor ($x-y$ coordinates and orientation). For the PointGoal task, the odometry sensor is used to compute the relative distance and angle of the pointgoal at each time step. The action space consists of 3 actions: \texttt{move\_forward} (0.25$m$), \texttt{turn\_left} ($10\degree$), \texttt{turn\_right} ($10\degree$).

\subsection{Navigation Policy}
\label{sec:training_details}
The source domain navigation policy ($\pi_s$) is trained using reinforcement learning. The reward for the PointGoal task is the decrease in geodesic distance to the point goal, and for the Exploration task is the increase in the explored area. The navigation policy is decomposed into a Policy Feature Extractor ($F_\pi$) and an Action Policy ($A_\pi$) as shown in Fig~\ref{fig:policy_decomposition}. $F_\pi$ is based on the 18-layer ResNet \cite{he2015deep}. $F_\pi$ outputs 128-dimensional policy-related representations ${\bf r}_I$ given RGB images of shape $3 \times 256 \times 256$. $A_\pi$ is based on a 2-layer GRU, which takes ${\bf r}_I$ along with relative distance and angle of the point goal (in the PointGoal task) or readings from the odometry sensor (in the Exploration task) as input. We train a separate policy for each task in the source Gibson domain using 72 scenes in the training set provided by~\cite{savva2019habitat}. The policies are trained using PPO~\cite{schulman2017proximal} for 30 million frames. In the PointGoal task, the agent stops when the relative distance of the PointGoal is less than $0.2m$, whereas, in the Exploration task, the agent explores for the maximum episode length of 500 steps. More architecture and hyperparamater details for navigation policy training are provided in supplementary.

\subsection{Implementation Details} Given a navigation policy, we use the proposed method for domain adaptation from Gibson to Replica and vice versa. The image encoders and decoders used in \methodabbr{} consist of several convolutional layers and residual blocks. The exact architecture is described in supplementary. Regarding hyperparameters, we use $\lambda_{im\_rec} = 10, \lambda_{lat\_rec} = 1, \lambda_{adv} = 1$ for all experiments. We scale $\lambda_{pol}$ based on the mean of image features for each task: $\lambda_{pol} = \frac{1}{\mathbb{E}_{I^s \sim \mathcal{S}^s}[||E_\pi(I^s)||_1]}$,
where $\mathbb{E}_{I^s \sim \mathcal{S}^s}[||E_\pi(I^s)||_1]$ is estimated using the Gibson images from the domain translation dataset.  We use the Adam optimizer \cite{kingma2014adam} with $0.0001$ initial learning rate, $\beta_1=0.5$ and $\beta_2=0.999$. The learning rate is halved every 100K iterations. We train the proposed model and all the baselines for 500K iterations.

\vspace{-2mm}
\subsection{Baselines}
\label{subsec:baselines}
\vspace{-4pt}
We compare the proposed \methodabbr ~model against the following baselines for domain adaptation: \\
1) \textbf{Direct Transfer:} This is the most common method of transferring navigation policies across domains, which involves directly testing the policy in the target domain without any fine-tuning.\\
2) \textbf{CycleGAN}~\cite{zhu2017unpaired} is a competitive and popular unsupervised image translation method. This method is designed for static image translation and is agnostic to the navigation policy.\\
3) \textbf{\methodabbr~w.o. Policy Loss:} This method is an ablation of the proposed method without the policy-based consistency loss. This ablation is conceptually very similar to prior works based on style and content disentanglement such as~\cite{MUNIT, lee2018diverse}. We use the architecture and hyperparameters from the \methodabbr{} model for this ablation to quantify the effect of the policy-based consistency loss on performance. 

\vspace{-2mm}
\section{Results}
\label{sec:results}

%We transfer the trained RL policies to the target domain using the proposed model, \methodname $\,$(\methodabbr) against 3 baselines defined in Section \ref{subsec:baselines}. We first present domain adaptation results from Gibson to Replica domains within the Habitat simulator and then present results of sim-to-real transfer from Gibson to real-world office scenes.

\setlength{\tabcolsep}{8pt}
\vspace{-3mm}
\begin{table}[t!]
\centering
\caption{\small \textbf{Results.} Comparisons between the proposed \methodname~(\methodabbr) and baselines on the PointGoal and Exploration tasks when transferred from the Gibson to the Replica domain.}
\label{tab:pg}
\footnotesize
% \vspace{-1mm}
\begin{tabular}{@{}llccccc@{}}
\toprule
                      &  & \multicolumn{2}{c}{PointGoal}   &           & \multicolumn{2}{c}{Exploration} \\ \midrule
                      &  & SPL            & Success Rate   &           & Explored Ratio  & Explored Area \\ \midrule
Direct Transfer       &  & 0.505          & 0.688          &           & 0.832           & 22.9          \\
CycleGAN              &  & 0.605          & 0.803          &           & 0.868           & 24.3          \\
\methodabbr{}~w.o. Policy Loss &  & 0.669          & 0.852          &           & 0.879           & 24.6          \\
\methodabbr{}                  &  & \textbf{0.712} & \textbf{0.881} & \textbf{} & \textbf{0.897}  & \textbf{25.3} \\ \bottomrule
\end{tabular}
\vspace{-4mm}
\end{table}

\subsection{Gibson to Replica}
\vspace{-6pt}
For evaluation in Replica, we generate a separate test set of 900 episodes (50 episodes in each of the 18 scenes) for both the PointGoal and Exploration tasks. The episodes are sampled using random starting locations and aggressive rejection sampling of near-straight-line episodes as in prior work~\cite{savva2019habitat,ans}. The performances of our method and all the baselines for both the tasks are presented in Table~\ref{tab:pg}. \methodabbr~outperforms all the baselines on both the tasks. It improves the SPL from 0.605 to 0.712 for PointGoal and Explored Ratio from 0.868 to 0.897 for Exploration as compared to the CycleGAN baseline. There's a considerable difference between \methodabbr{} and the ablation, indicating the importance of policy-based consistency loss. PBIT performed better than all the baselines consistently across 3 different runs with a small standard deviation of 0.007 SPL and 0.004 Explored Ratio. The Explored Ratios of all the methods in Table~\ref{tab:pg} are high on an absolute level because the Replica scenes are relatively small usually having one or two rooms. Just turning on the spot leads to an explored ratio of 0.436.

\begin{figure}[t]
    \vspace{-3mm}
    \centering
    \includegraphics[width=0.99\linewidth, angle=0]{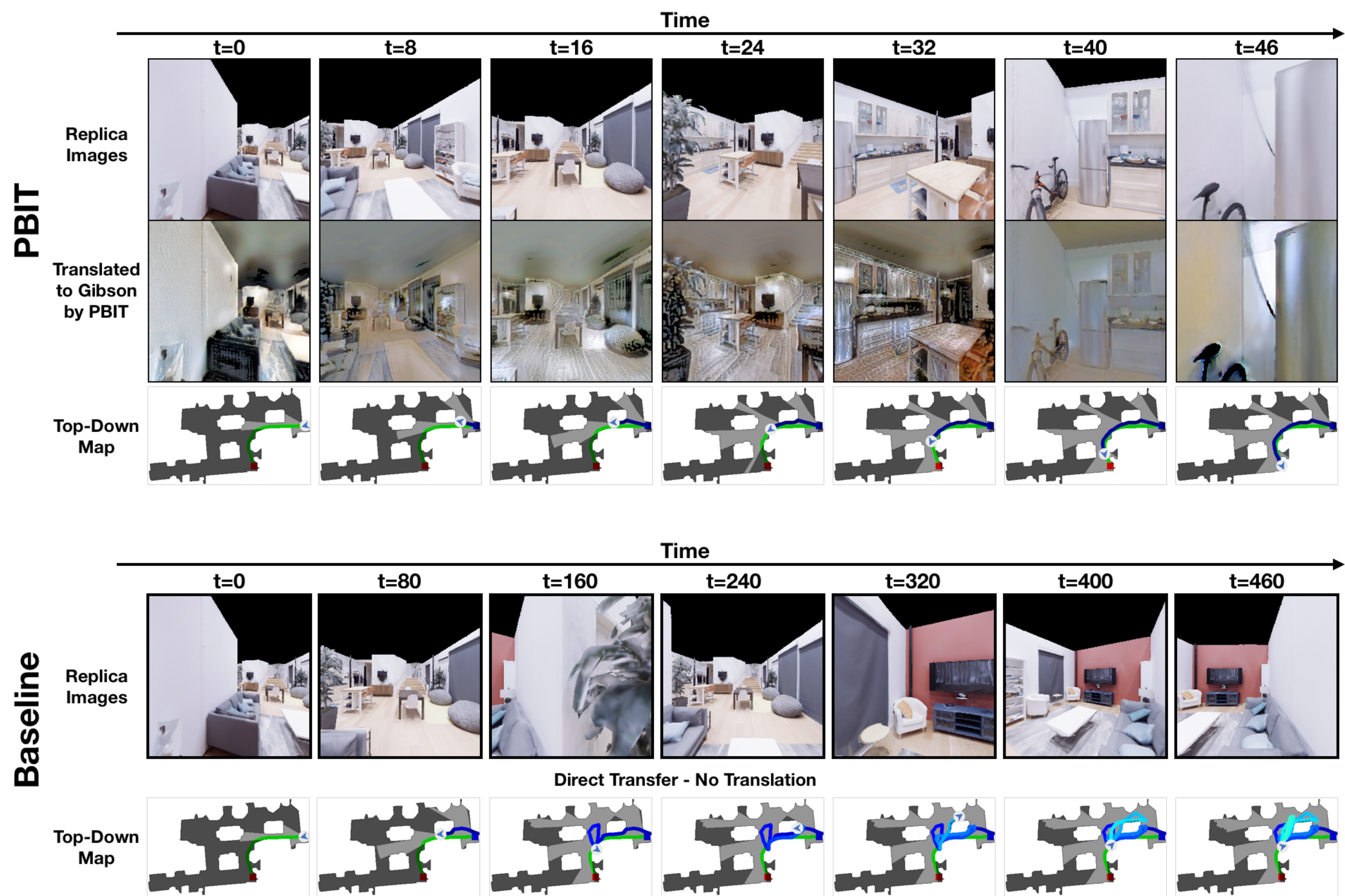}
    % \vspace{-1mm}
    \caption{\small \textbf{Trajectory Comparison between \methodabbr{} and the Direct Transfer Baseline on the PointGoal Task in Replica. } 
    The upper half of the figure is the trajectory of \methodabbr{}: the agent successfully navigates from a corner of the apartment to the fridge in 46 steps, by seeing the translated images by \methodabbr{}. The agent takes almost the shortest path possible, as shown in the Top-Down Map (not visible to the agent). The lower half of the figure is the trajectory of the Direct Transfer baseline on the same test episode. The Direct Transfer agent fails to navigate to the target location and gets lost, even after 460 steps.  
    }
    \label{fig:replica_trajectory}
    \vspace{-5mm}
\end{figure}

In Figure~\ref{fig:replica_trajectory}, we visualize an example trajectory for the PointGoal task using the proposed method ~\methodabbr~(Fig~\ref{fig:replica_trajectory} above) and the Direct Transfer baseline (Fig~\ref{fig:replica_trajectory} below) for the same episode specification. The figure shows the images observed by the agent in the target domain, the translated images, and a top-down map (not visible to the agent) showing the point goal coordinates and the agent's path. The PBIT model successfully reaches the PointGoal within 46 steps while the Direct Transfer baseline is unable to reach the goal. The image translations indicate that policy-relevant characteristics of the image such as corners of obstacles, walls and free space are preserved during the translation.

\iffalse
\begin{figure}[t]
    \centering
    \includegraphics[width=0.99\linewidth, angle=0]{images/realworld_4by8.pdf}
    % \vspace{-3mm}
    \caption{\small \textbf{Real World Images Translated to Gibson.} 
    The first and fifth columns are the input images from Real World during test time. The other columns are images translated to
    Gibson domain by \methodabbr{}. Although the agent has never been trained to navigate with Real World images, it can recognize the translated images (which preserve the policy-relevant features) and perform well on PointGoal in Real World.
    }
    \label{fig:realworld_examples}
\end{figure}
\fi

\vspace{-1mm}
\subsection{Gibson to Real-world}
\vspace{-6pt}
For the real-world experiments, we conduct 18 trials each for the proposed method and the Direct Transfer baseline for the PointGoal task.  We transfer the navigation policy to a LoCoBot~\cite{locobot} using the PyRobot API~\cite{pyrobot2019} for both the methods. We conduct trials in 2 scenes, Seen and Unseen. 151 images among the set of the 1125 real-world images used for training the \methodabbr{} model were sampled in the Seen scene, whereas none of the images were sampled from the Unseen scene. We terminate the episode if the robot collides with an obstacle and count it as a failure. We allow a maximum episode length of 99 steps in the real-world experiments.

Each trial specification and the corresponding results are presented in Table \ref{table:real_world}. \methodabbr~achieves an absolute improvement of $55\%$ ($70\%$ vs $15\%$) in success rate over the Direct Transfer baseline across all the trials. \methodabbr~also has a much lower collision rate as compared to the baseline. Surprisingly, the PBIT model achieves 100\% success rate in the Seen scene, achieving an absolute improvement of $78.8\%$ ($100\%$ vs $22.2\%$). These results indicate that the navigation policy can be reliably transferred to the real-world using the proposed method given access to a few images in the real-world scene. Even in the unseen scene, PBIT leads to a large absolute improvement of $22.3\%$ ($55.6\%$ vs $33.3\%$) over the Direct Transfer baseline.

In Figure~\ref{fig:realworld_trajectory}, we show an example of a successful trajectory in the Unseen real-world scene using~\methodabbr. It shows some of the images seen by the agent during the trajectory, the corresponding translations, and a third-person view of the robot. The trajectory shows the~\methodabbr{} is able to successfully navigate around the blue chair obstacle to reach the pointgoal. The image translations shown in the figure also indicate that the model generates good translations similar to images in the Gibson domain. For example, the dark grey carpet floors in the office space scenes in the real-world are successfully translated to brown floors, representative of wooden floors of apartment scenes in the Gibson domain. Similarly, wooden doors and staircases in the real-world are translated to off-white walls which are common in Gibson scenes. At the same time, navigation relevant details such as the boundary between the floor and walls and other obstacles, are preserved during translation.

\begin{figure}[]
    \centering
    \vspace{-3mm}
    \includegraphics[width=0.95\linewidth, angle=0]{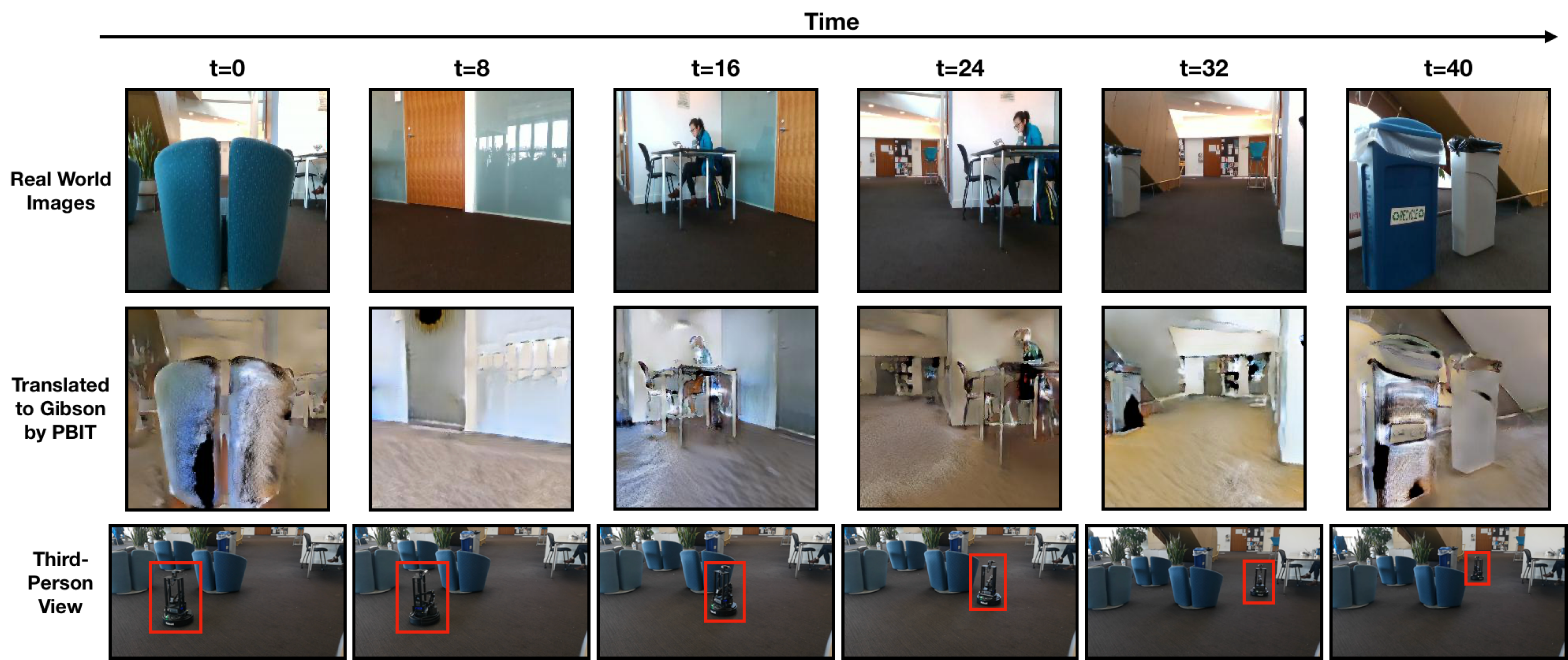}
    \vspace{-1mm}
    \caption{\small \textbf{Sample Real World Trajectory on PointGoal Task.} 
    Figure contains raw inputs from Real World (row one), translated Gibson images by PBIT (row 2), and a third-person perspective from the back. The PBIT agent successfully reached it's destination (a trash can) by avoiding an obstacle (a chair) in its way.
    }
    \label{fig:realworld_trajectory}
    \vspace{-2mm}
\end{figure}

\setlength{\tabcolsep}{4.3pt}
\begin{table}[h!]
	\centering
	\caption{\small \textbf{Real-world results.} Table comparing the performance of the Direct Transfer baseline and the proposed method \methodabbr{} across 18 trials in two scenes in the real-world. The training set for PBIT consists of some images sampled from Scene 1: Corridors (Seen), but no image from Scene 2: Public kitchen area (Unseen). The episode specification (starting relative distance and angle to the PointGoal) are shown in the left columns. Identical starting locations and episode specifications were used to evaluate both the methods. The performance of the baseline and \methodabbr{} are shown in the center and right, respectively. Success is abbreviated as Succ.}
\label{table:real_world}
% 	\vspace{-1mm}
% \vskip - 0.1in
%\resizebox{\textwidth}{!}{%
	{\small
\begin{tabular}{@{}ccccccccccccc@{}}
\toprule
\multicolumn{3}{c}{Episode Specification}                                          &  & \multicolumn{4}{c}{Baseline: Direct Transfer}                                              &  & \multicolumn{4}{c}{PBIT}                                                                   \\ \midrule
	Ep No & Dist ($m$) & Angle ($\degree$)  &  & Steps & Collision & \begin{tabular}[c]{@{}c@{}}Final\\Dist\end{tabular} & Succ. &  & Steps & Collision. & \begin{tabular}[c]{@{}c@{}}Final\\Dist\end{tabular} & Succ. \\ \midrule\\[-0.7em]
%\multicolumn{13}{l}{\textbf{Scene 1: Wooden corridor with intense ground reflection (Not in training set of PBIT)}}\\
%1     & 4.00 & 0.00   & 0                                                      &  & 99    & 0     & 5.62                                                         & 0   &  & 99    & 0     & 3.48                                                         & 0   \\
%2     & 2.83 & 45.00  & 0                                                      &  & 68    & 1      & 4.12                                                         & 0   &  & 99    & 0     & 1.80                                                         & 0   \\ 
%SCENE AVG    & - & -  & - &  & 99 & 50\%     & 4.87 & 0\%   &  & \textbf{99}    & \textbf{0\%}   &  \textbf{2.64}  & 0\%
%\\ \\
	\multicolumn{13}{l}{\textbf{Scene 1 (Seen): Corridors}}\\\\[-0.7em]

1    & 2.83 & 45.00                                                        &  & 99    & 0     & 3.41                                                         & 0   &  & 21    & 0     & 0.15                                                         & 1    \\
2    & 4.47 & 333.43                                                       &  & 99    & 0     & 4.68                                                         & 0   &  & 41    & 0     & 0.02                                                         & 1    \\
3    & 4.00 & 0.00                                                         &  & 99    & 0     & 4.43                                                         & 0   &  & 41    & 0     & 0.13                                                         & 1    \\
4    & 5.39 & 21.80                                                        &  & 10    & 1      & 4.60                                                         & 0   &  & 32    & 0     & 0.13                                                         & 1    \\
5    & 4.24 & 315.00                                                      &  & 99    & 0     & 6.20                                                         & 0   &  & 73    & 0     & 0.14                                                         & 1    \\
6    & 4.00 & 0.00                                                         &  & 36    & 1      & 3.88                                                         & 0   &  & 63    & 0     & 0.09                                                         & 1    \\
7    & 1.41 & 45.00                                                        &  & 99    & 0     & 1.95                                                         & 0   &  & 9     & 0     & 0.19                                                         & 1    \\
8    & 1.41 & 135.00                                                       &  & 50    & 0     & 0.15                                                         & 1    &  & 29    & 0     & 0.07                                                         & 1    \\
9    & 4.47 & 26.57                                                        &  & 99    & 0     & 4.73                                                         & 0   &  & 27    & 0     & 0.19                                                         & 1    \\ \\[-0.7em]
	\multicolumn{3}{l}{Scene Avg}  &  & 92 & 22.2\%     & 3.78 & 11.1\%   &  & \textbf{37.3}    & \textbf{0\%}   &  \textbf{0.13}  & \textbf{100\%} \\\\[-0.7em]
	\midrule\\[-0.7em]

	\multicolumn{13}{l}{\textbf{Scene 2 (Unseen): Public kitchen area}}\\\\[-0.7em]

1     & 2.00 & 0.00                                                        &  & 32    & 1      & 0.50                                                         & 0   &  & 36    & 0     & 0.18                                                         & 1    \\
2     & 2.00 & 0.00                                                        &  & 35    & 1      & 0.56                                                         & 0   &  & 10    & 0     & 0.04                                                         & 1    \\
3     & 2.24 & 333.43                                                      &  & 31    & 0     & 0.02                                                         & 1    &  & 16    & 0     & 0.07                                                         & 1    \\
4     & 2.24 & 153.43                                                      &  & 38    & 0     & 0.08                                                         & 1    &  & 43    & 0     & 0.09                                                         & 1    \\
5     & 4.12 & 345.96                                                       &  & 80    & 1      & 4.10                                                         & 0   &  & 44    & 1      & 1.96                                                         & 0   \\
6     & 4.47 & 26.57                                                        &  & 99    & 0     & 4.01                                                         & 0   &  & 99    & 0     & 2.85                                                         & 0   \\
7     & 4.47 & 26.57                                                        &  & 99    & 0     & 4.49                                                         & 0   &  & 70    & 0     & 0.14                                                         & 1    \\
8    & 5.39 & 21.80                                                        &  & 99    & 0     & 5.50                                                         & 0   &  & 99    & 0     & 2.73                                                         & 0   \\
9    & 2.83 & 45.00                                                        &  & 99    & 0     & 3.57                                                         & 0   &  & 99    & 0     & 2.84                                                         & 0   \\\\[-0.7em]
	\multicolumn{3}{l}{Scene Avg} &  & 77.5 & 33.3\%     & 2.54 & 22.2\%   &  & \textbf{59.0}    & \textbf{11.1\%}   &  \textbf{1.21}  & \textbf{55.6\%} \\ \\[-0.7em]
\midrule
\multicolumn{3}{l}{Overall} &  & 84.8 & 27.7\%     & 3.16                                                         & 16.7\%   &  & \textbf{48.2}    & \textbf{5.6\%}   &  \textbf{0.67}  & \textbf{77.8\%}  \\ \bottomrule
\end{tabular}
	}
\vspace{-4mm}
\end{table}

\iffalse
\begin{minipage}{\linewidth}
%\begin{figure}[t]
    \centering
    \includegraphics[width=0.9\linewidth, angle=0]{images/realworld_4by8.pdf}
    \vspace{-3mm}
    \captionof{figure}{\small \textbf{Real World Images Translated to Gibson.} 
    The first and fifth columns are the input images from Real World during test time. The other columns are images translated to
    Gibson domain by \methodabbr{}. Although the agent has never been trained to navigate with Real World images, it can recognize the translated images (which preserve the policy-relevant features) and perform well on PointGoal in Real World.
    }
    \label{fig:realworld_examples}
%\end{figure}

%\begin{figure}[t]
    \centering
    \includegraphics[width=0.9\linewidth, angle=0]{images/realworld_traj.pdf}
    \vspace{-3mm}
    \captionof{figure}{\small \textbf{Sample Real World Trajectory on PointGoal Task.} 
    Figure contains raw inputs from Real World (row one), translated Gibson images by PBIT (row 2), and a third-person perspective from the back. The PBIT agent successfully reached it's destination (a trash can) by avoiding an obstacle (a chair) in its way.
    }
    \label{fig:realworld_trajectory}
%\end{figure}
\end{minipage}
\fi

%\begin{minipage}{\linewidth}

%\end{minipage}

\vspace{-4pt}
\section{Conclusion}
\label{sec:conclusion}
\vspace{-6pt}
In this paper, we proposed a domain adaptation method for transferring navigation policies from simulation to the real-world. Given a navigation policy in the source domain, our method translates images from the target domain to the source domain such that the translations are consistent with the task-specific and domain-invariant representations learnt by the given policy. Our experiments across two different tasks for domain transfer in simulation show that the proposed method can improve the performance on the transferred navigation policies over baselines. We also show strong performance of navigation policies transferred from simulation to the real-world using our method. In this paper, we considered navigation tasks involving mostly spatial reasoning. In the future, the proposed method can be extended to navigation tasks involving more geometric reasoning by incorporating semantic consistency losses along with the policy consistency losses.

%\FloatBarrier

%\section*{Broader Impact}
%In this work, we study the domain adaptation from simulation to real-world and from real-world to simulation. We should be aware of the malicious usage for our framework. For instance, people with bad intentions can first build a simulated environment for a highly guarded place, then training an agent to find the simulation's loopholes. The loopholes found in the simulation can potentially reflect the ones that appear in reality.

{
\small
\bibliography{example}
\bibliographystyle{plain}
}
 
\appendix

\clearpage
\section{Visualization of Policy Representations}
\begin{figure}[h]
    \centering
    \includegraphics[width=0.9\textwidth, angle=0]{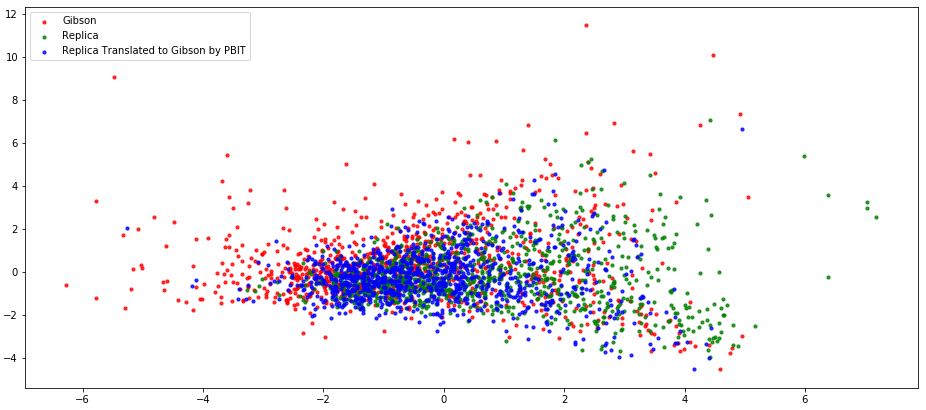}
    \vspace{-3mm}
    \caption{We use PCA to visualize and compare the 128-dimensional task-specific feature vectors produced by PointGoal agent trained in Gibson, when given Gibson images, Replica images, or Replica images translated to Gibson by PBIT as input. The figure shows the translated images by PBIT bridge the policy domain gap between Gibson and Replica.
    }
    \label{fig:pca}

    \includegraphics[width=0.9\textwidth, angle=0]{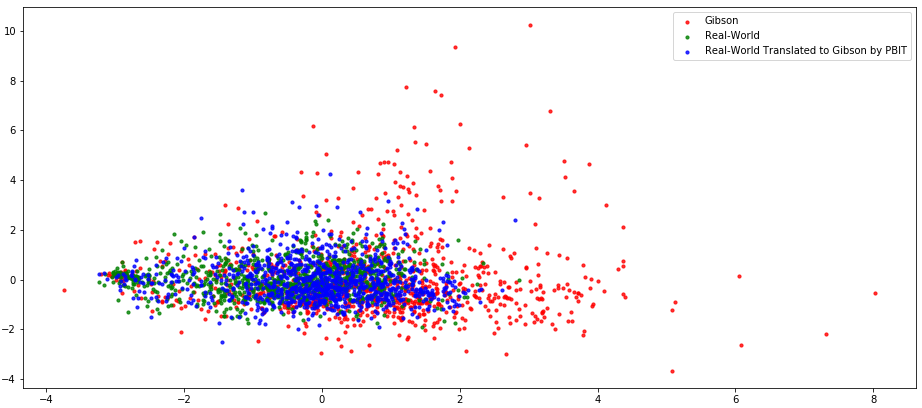}
    \vspace{-3mm}
    \caption{We use PCA to visualize and compare the 128-dimensional task-specific feature vectors produced by PointGoal agent trained in Gibson, when given Gibson images, Real-World images, or Real-World images translated to Gibson by PBIT as input. The figure shows the translated images by PBIT bridge the policy domain gap between Gibson and Real-World.
    }
    \label{fig:pca_real}
\end{figure}

We analyze the policy representation before and after translation by reducing the dimensionality of the policy representations using Principle Component Analysis (PCA)~\cite{wold1987principal}. In Figures~\ref{fig:pca} and~\ref{fig:pca_real}, we visualize the policy representations reduced to 2 dimensions using PCA in Replica and Real-World respectively. Both figures show that PBIT brings the representations of target domain Replica/Real-World images closer to the distribution of representations of Gibson images.

%% ADD FIGURES HERE

\clearpage

\section{Navigation Policy Training Details}
\begin{figure}[t]
    \centering
    \includegraphics[width=0.99\textwidth, angle=0]{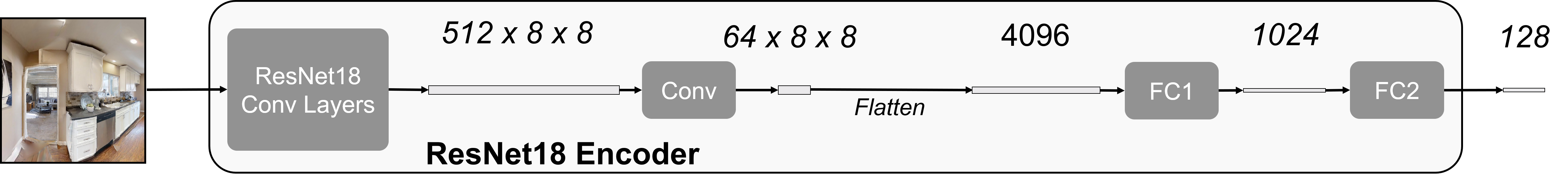}
    \caption{An illustration of the network architecture of the Policy Feature Extractor.}
    \label{fig:agent_archi}
\end{figure}

\textbf{Policy Architecture.} The navigation policy is decomposed into a Policy Feature Extractor ($F_\pi$) and a Action Policy ($A_\pi$) as shown in Fig~\ref{fig:policy_decomposition}. $F_\pi$ is based on the 18-layer ResNet \cite{he2015deep}. The architecture of $F_\pi$ is shown in Figure~\ref{fig:agent_archi}. $F_\pi$ outputs 128-dimensional policy-related representations ${\bf r}_I$ given RGB images of shape $3 \times 256 \times 256$. $A_\pi$ is based on a 2-layer GRU, which takes ${\bf r}_I$ together with readings from either the GPS+Compass sensor in PointGoal task or the base odometry sensor in Exploration task. 

% #with 128-dimensional hidden state, which takes in 

% % , where we 
% % remove the last two layers (average pooling layer and fully connected layer) from the 18-layer ResNet, and add a convolution layer of 64 output channels, kernel size 1 and stride 1, followed by two fully connected layers of dimension 1024 and 128 respectively. All the added layers have ReLU activation. We also use Dropout \cite{JMLR:v15:srivastava14a} on the fully connected layer of dimension 1024. 

% Hence the visual encoder takes in RGB images of shape $3 \times 256 \times 256$ and outputs policy-related representations of shape $128$. The policy encoder $A_\pi$ is parameterized by a 2-layer GRU with 128-dimensional hidden state. The input to the 2-layer GRU is the 128-dimensional policy-related representations produced by the visual encoder, concatenated with either the 2-dimensional input from the GPS+Compass sensor in PointGoal task, or the 3-dimensional input from the base odometry sensor in Exploration task. The output from the 2-layer GRU is then used to output a softmax distribution over the action space as well as an estimate of the value function. 

%\vspace{-2mm}
\textbf{Policy Training.} The navigation policies are trained using PPO \cite{schulman2017proximal} with Generalized Advantage Estimation \cite{schulman2015highdimensional}. The training dataset contains episodes in 72 training scenes in Gibson as provided by \cite{savva2019habitat}. We use 8 concurrent workers and train for a total of 30 million frames. The time horizon for a PPO update is 128 steps, number of epochs per PPO update is 2, and clipping parameter is set to 0.2. We use discount factor 0.99, GAE parameter 0.95, and the Adam optimizer \cite{kingma2014adam} with learning rate $2.5 \times 10^{-4}$.

\section{PBIT Model Architecture Details}

We use several convolutional layers and residual blocks to construct the image encoders $E_I^s, E_I^t$ and image decoders $D_I^s, D_I^t$. We use Instance Normalization \cite{Ulyanov_2017_CVPR} in $E_I^s, E_I^t$
and Adaptive Instance Normalization \cite{Huang_2017_ICCV} in $D_I^s, D_I^t$. For the discriminators ${\rm Dis}^{s}, {\rm Dis}^{st}$, we adopt the multi-scale discriminator architecture proposed by \cite{Wang_2018_CVPR}. Detailed descriptions of the architecture are given in Table~\ref{tab:pbit_arch}. %For the CycleGAN baseline, we use the architecture proposed in the CycleGAN paper \cite{zhu2017unpaired}. 

\begin{table}[h]
    \caption{Architecture specification of different parts of the proposed PBIT model.}
    \label{tab:pbit_arch}
    \begin{minipage}{.245\linewidth}
    \begin{subtable}[t]{1in}
      \caption{Style Encoder}
      \centering
        \begin{tabular}{c}
        \hline \hline
        % \vspace{0.1in}
        $(256\times 256)$ RGB\\
        \hline
        Conv(32,(7x7),1)\\
        \hline
        Conv(64,(4x4),2)\\
        \hline
        Conv(128,(4x4),2)\\
        \hline
        Conv(128,(4x4),2)\\
        \hline
        Conv(128,(4x4),2)\\
        \hline
        global\_avg\_pool\\
        \hline
        dense(8)\\
        \hline
        $\rightarrow s\in \mathbb{R}^{8}$\\
        \hline \hline
        \end{tabular}
    \end{subtable}
    \end{minipage}%
    \begin{minipage}{.25\linewidth}
    \begin{subtable}[t]{1in}
      \caption{Content Encoder}
      \centering
        \begin{tabular}{c}
        \hline \hline
        % \vspace{0.1in}
        $(256\times 256)$ RGB\\
        \hline
        Conv(32,(7x7),1)\\
        \hline
        Conv(64,(4x4),2)\\
        \hline
        Conv(128,(4x4),2)\\
        \hline
        Residual(128)\\
        \hline
        Residual(128)\\
        \hline
        Residual(128)\\
        \hline
        Residual(128)\\
        \hline
        $\rightarrow c\in \mathbb{R}^{128\times 64\times 64}$\\
        \hline \hline
        \end{tabular}
    \end{subtable}
    \end{minipage}%
    \begin{minipage}{.25\linewidth}
    \begin{subtable}[t]{1in}
      \caption{Decoder}
      \centering
        \begin{tabular}{c}
        \hline \hline
        $c$ and AdaIN$(s)$\\
        \hline
        Residual(128)\\
        \hline
        Residual(128)\\
        \hline
        Residual(128)\\
        \hline
        Residual(128)\\
        \hline
        UpSampling(2x2)\\
        \hline
        Conv(64,(5x5),1)\\
        \hline
        UpSampling(2x2)\\
        \hline
        Conv(32,(5x5),1)\\
        \hline
        UpSampling(2x2)\\
        \hline
        Conv(3,(7x7),1)\\
        \hline
        $(256\times 256)$ RGB\\
        \hline \hline
        \end{tabular}
    \end{subtable}
    \end{minipage}%
    \begin{minipage}{.25\linewidth}
    \begin{subtable}[t]{1in}
      \caption{Discriminator}
      \centering
        \begin{tabular}{c}
        \hline \hline
        % \vspace{0.1in}
        Multi-scale Input\\
        \hline
        Conv(64,(4x4),2)\\
        \hline
        Conv(128,(4x4),2)\\
        \hline
        Conv(256,(4x4),2)\\
        \hline
        Conv(512,(4x4),2)\\
        \hline \hline
        \end{tabular}
    \end{subtable}
    \end{minipage}%
\end{table}

% \usepackage{booktabs}
\iffalse
\begin{table}[]
\centering
\caption{PPO Hyperparameters}
\label{table:ppo_hyperparameter}
\setlength{\tabcolsep}{20pt}
\begin{tabular}{@{}ll@{}}
\toprule
Hyperparameter     & Value          \\ \midrule
Clipping parameter & 0.2            \\
Horizon (steps per update)     & 128            \\
Adam stepsize      & $2.5 \times 10^{-4}$         \\
Num. PPO epochs        & 2              \\
Minibatch size     & $128 \times 2$ \\
Discount factor           & 0.99           \\
GAE parameter      & 0.95           \\
Value Loss coeff.  & 0.5            \\
Entropy coeff.     & 0.01           \\ \bottomrule
\end{tabular}
\end{table}
\fi

\clearpage
\section{Additional Trajectory Visualizations}
\begin{figure}[h]
\centering
\includegraphics[width=0.99\textwidth, angle=0]{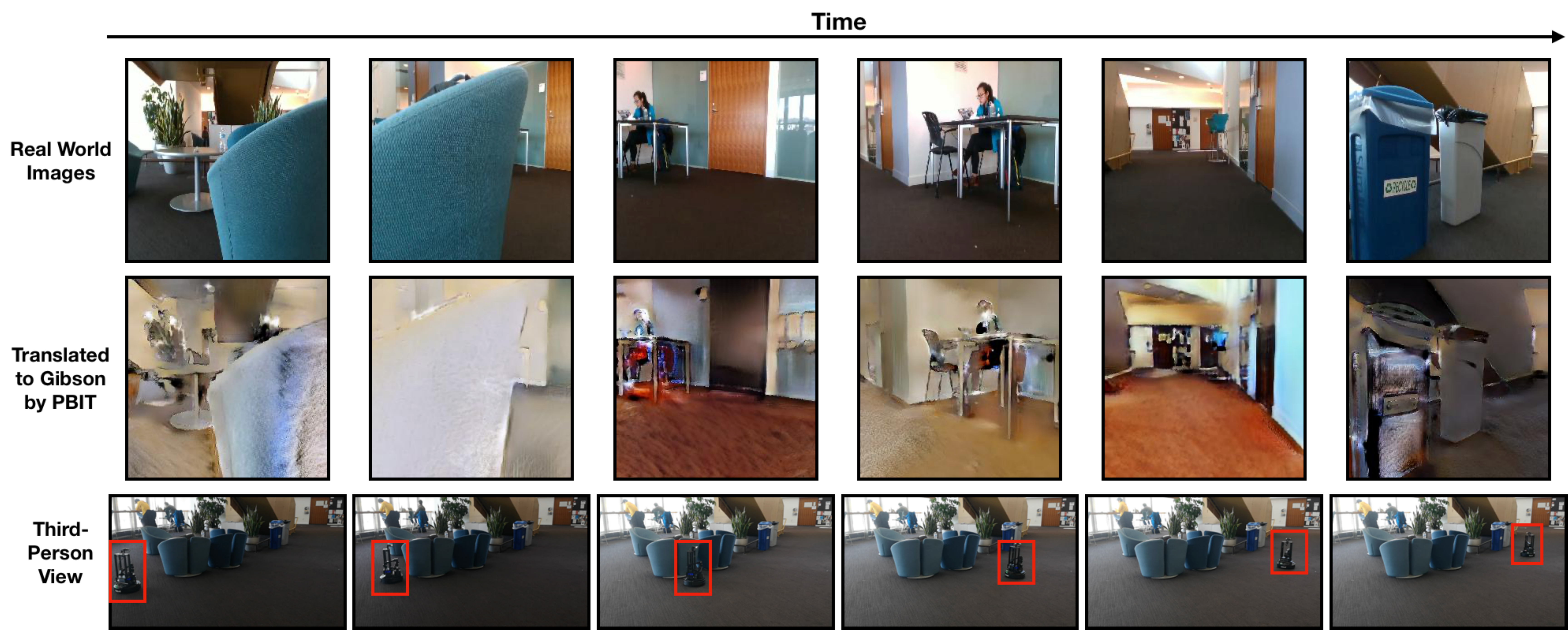}

\includegraphics[width=0.99\textwidth, angle=0]{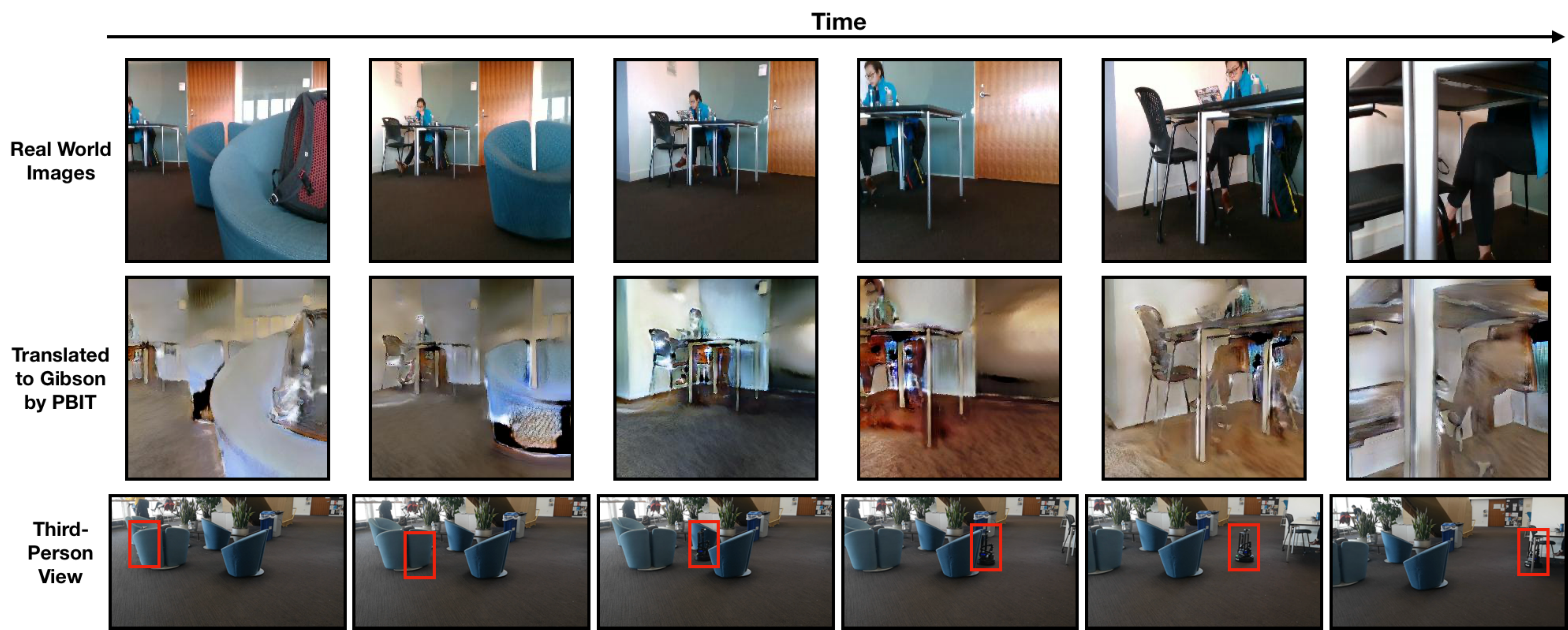}
\caption{\small \textbf{Additional Real World Trajectories.} 
    Figure showing raw inputs, translated Gibson images by PBIT, and a third-person perspective for two trajectories in the real-world for the PointGoal task.. 
    }
\end{figure}

\begin{figure}[]
\centering
\includegraphics[width=0.99\textwidth, angle=0]{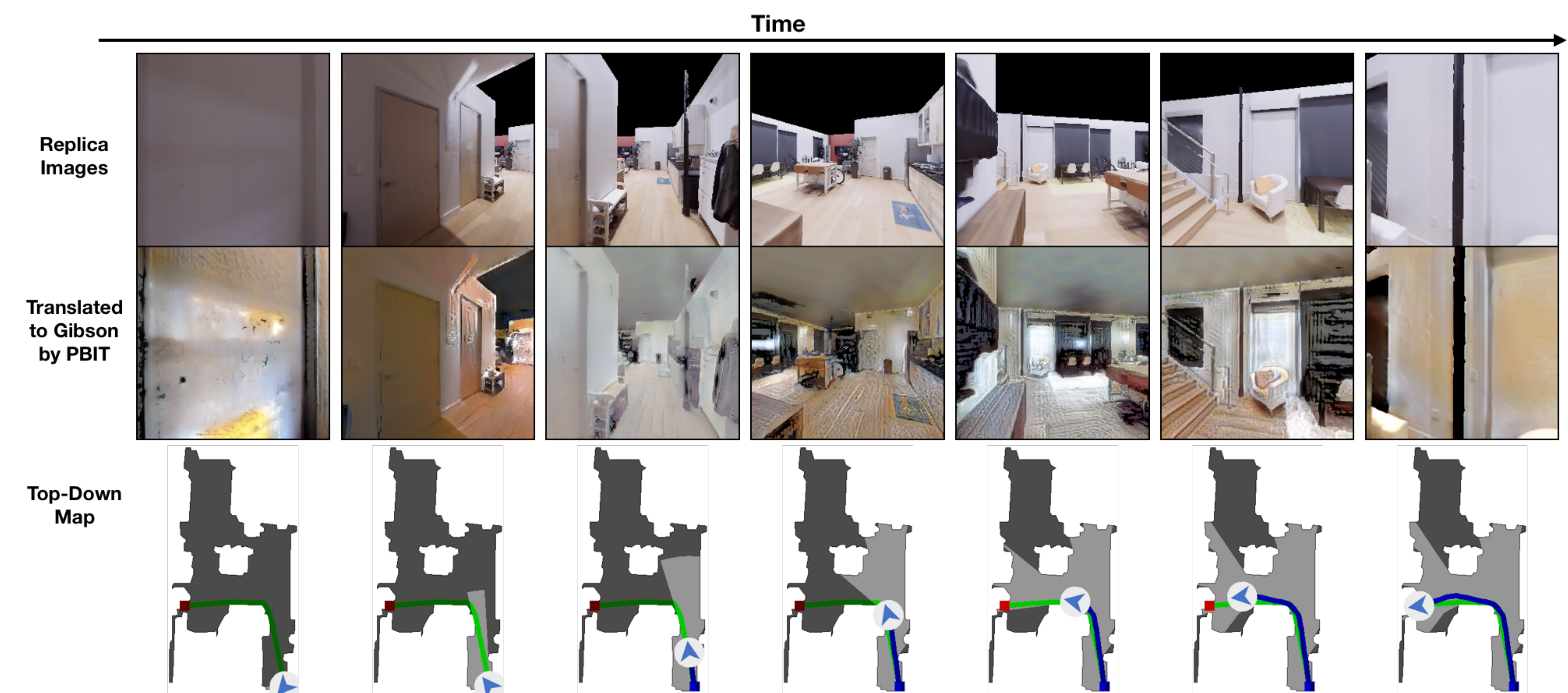}

\includegraphics[width=0.99\textwidth, angle=0]{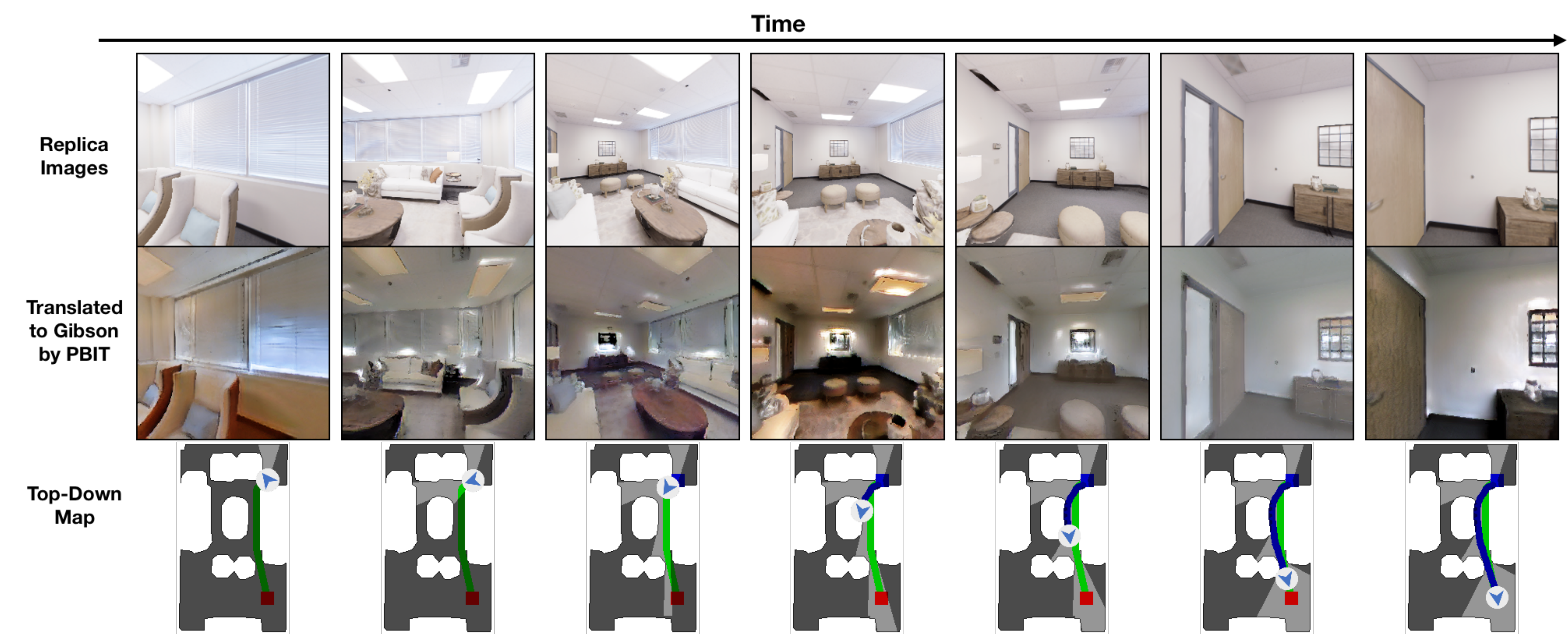}
 
\includegraphics[width=0.99\textwidth, angle=0]{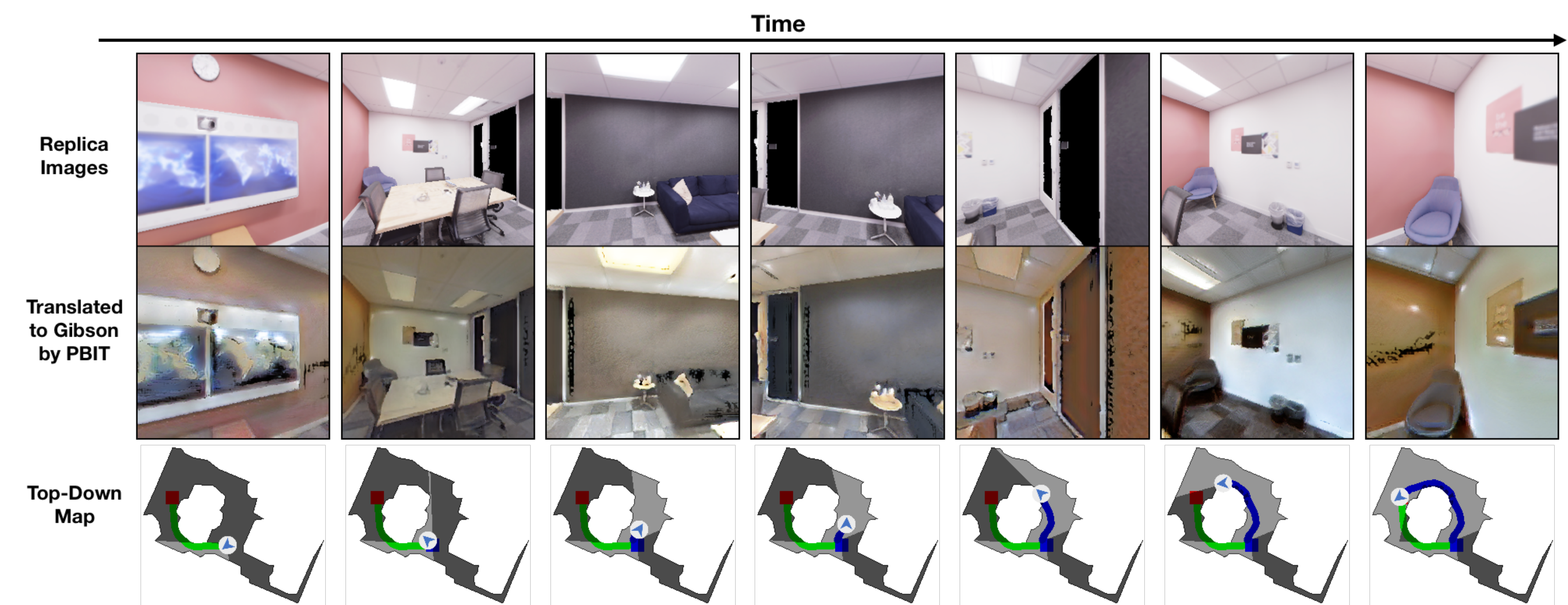}
\caption{\small \textbf{Additional Replica Trajectories.} 
    Figure showing raw inputs, translated Gibson images by PBIT, and the ground-truth top-down map (not visible to the agent) for three trajectories in the Replica domain. 
    }
\end{figure}

\iffalse
\begin{figure}[]
\centering

\includegraphics[width=0.9\textwidth, angle=0]{supp/sim_traj_6.pdf}

\includegraphics[width=0.9\textwidth, angle=0]{supp/sim_traj_2.pdf}

\includegraphics[width=0.9\textwidth, angle=0]{supp/sim_traj_7.pdf}

\includegraphics[width=0.9\textwidth, angle=0]{supp/sim_traj_3.pdf}

\end{figure}
\fi
%\subsection{Real World Experiment} 
%\label{sec:real_world_exp}
%\subsubsection*{Robot configuration.} We present results of unsupervised zero-shot simulation to real-world indoor navigation on a LoCoBot \cite{locobot}. The robot has an RGB camera 60cm from the ground, and is programmed to take action space: [stop, forward $0.25m$, left $10^\circ$, right $10^\circ$]. %The robot is very slow, like a sloth.

%\vspace{-2mm}
%\subsubsection*{Data collection and training.} By taking pictures at random locations with the robot, we collect a total of 1125 images from three indoor locations: 1) 397 images from a meeting room with chairs and tables. 2) 728 images from the corridor of a building. 3) 151 images from a large study place chairs and tables. 

%We train an agent of camera height $60cm$ in Gibson with the same architecture as in section \ref{sec:training_details}, and a policy based transfer model from 7500 random Gibson images and the 1125 real-world images.

\end{document}